\documentclass[10pt,a4paper]{article}
\usepackage[english]{babel}
\usepackage[T1]{fontenc} 
\usepackage{amssymb}
\usepackage{amsmath}
\usepackage{fancyhdr}
\usepackage{verbatim}
\usepackage{lastpage}
\usepackage{ifthen}
\usepackage{float}
\usepackage{multirow,booktabs}
\usepackage{lscape,longtable}
\usepackage{longtable}
\usepackage{tabu}
\usepackage[top = 0.7in, bottom = 0.7in, left = 0.7in, right = 0.7in]{geometry} 

\usepackage[pdftex]{color,graphicx}
\usepackage{hyperref}
\begin{document}
\title{Drivers' attention detection: a systematic literature review}


\author{Luiz G. Véras\thanks{Federal Institute of São Paulo, Bragança Paulista, São Paulo, Brazil (e-mail: gustavo\_veras@ifsp.edu.br). }, Anna K. F. Gomes\thanks{Federal Institute of São Paulo, Cubatão, São Paulo, Brazil (e-mail: anna.gomes@ifsp.edu.br). }, Guilherme A. R. Dominguez\thanks{the Federal Institute of São Paulo, Cubatão, São Paulo, Brazil (e-mail: g.rey@aluno.ifsp.edu.br). }  and Alexandre T. Oliveira\thanks{Federal Institute of São Paulo, Bragança Paulista, São Paulo, Brazil (e-mail: tomazati@ifsp.edu.br). }
}
\date{2021}
\medskip
\maketitle

\smallskip

\noindent{\small{{\textbf Abstract}
Countless traffic accidents often occur because of the inattention of the drivers. Many factors can contribute to distractions while driving, since objects or events to physiological conditions, as drowsiness and fatigue, do not allow the driver to stay attentive. The technological progress allowed the development and application of many solutions to detect the attention in real situations, promoting the interest of the scientific community in these last years. Commonly, these solutions identify the lack of attention and alert the driver, in order to help her/him to recover the attention, avoiding serious accidents and preserving lives. Our work presents a Systematic Literature Review (SLR) of the methods and criteria used to detect attention of drivers at the wheel, focusing on those methods based on images. As results, 50 studies were selected from the literature on drivers' attention detection, in which 22 contain solutions in the desired context. The results of SLR can be used as a resource in the preparation of new research projects in drivers' attention detection.
}}

\medskip



\section{Introduction}
The lack of attention while driving can lead to severe accidents in the traffic, which can involve people with serious injuries or even fatalities. According to the World Health Organization, nearly $1.35$ millions of people die every year due to traffic accidents, in which more than the half are pedestrians, cyclists and motorcyclists\footnote{https://www.who.int/news-room/fact-sheets/detail/road-traffic-injuries}. In particular, the Brazilian Association for Traffic Medicine (ABRAMET) related that the sleepiness is the major cause of the traffic accidents, around $42\%$ of the occurrences \cite{jorge2008acidentes}. The excessive consume of alcohol and fatigue can be related to the sleepiness while driving, and also young people and men present drowsiness leading to bigger risks to accidents \cite{stutts2003driver}.

The significant increase of accidents involving fatigue and drowsiness of drivers leads to a quick need to develop automatic and reliable systems to detect attention and fatigue, in order to emit an alert before a potentially dangerous situation may occur. These resources are denominated Advanced Driver Assistance System (ADAS) and coupled to vehicles can aid the drivers in many situations, as alerting the loss or lack of attention. 

To identify the reduction of attention, an ADAS can be supported in different attention criteria, such as the eye gaze, head position, alterations in the heartbeat rhythm or even in brain functions. Data related to these criteria can be obtained by cameras, cellphones, electrodes, special glasses, and many other types of sensors that can be coupled to the driver or vehicle. These data are subjected to computational methods to be analyzed and, thus, it becomes possible to identify the level of attention of the driver. The results of this analysis are informed to the ADAS, which can send an alert or not to the driver. 

There are a considerable variety of methods and criteria to be used in attention detection. It is important to know the solutions that already exist in this context, in order to not only avoid rework but also to find a more adequate approach for the desired application. To properly manage the search for this knowledge, it is adequate to employ a well defined methodology to perform the review of the existing methods in the literature. Based on this, the goal of this work is to describe the results of a Systematic Literature Review (SLR) of the computational methods used for drivers' attention detection.

SLR is a popular methodology to select primary studies in software engineering \cite{kitchenham2007guidelines,budgen2006performing} and medicine \cite{david2000evidence}. The following review is structured to initially define the problem reasoning that defines the research scope. Then, a \textit{review protocol} is specified, containing the guidelines to execute the review: the tools to search the studies; the terms to be searched; the inclusion and exclusion criteria; the data to be extracted. The protocol must be developed by a team of researchers to achieve a consensus about the SLR guidelines. These researchers support the review process, while a reviewer executes the protocol. The main advantage of SLR is the evaluation of the review process by third parties, thus reducing the bias of the reviewer \cite{kitchenham2007guidelines}.

The presented work is organized as follows. The basic concepts are introduced in Section~\ref{sec:crit_aval}. In Section~\ref{sec:trab_rela} we present other reviews related to drivers' attention detection.  The review protocol is described in details in Section~\ref{sec:prot_rev}. Section~\ref{sec:exec_rev} presents the execution process, the results obtained at each stage and the validation of the process. In Section~\ref{sec:resul_disc}, the selected primary studies are summarized and discussed.

\section{Attention Criteria}
\label{sec:crit_aval}

To start the discussion about attention detection, it is important to define some concepts in advance. The attention, according to \cite{michaelis2015dicionario}, is defined as the concentration of mental activity in a certain person or {thing}. Within the context of this work, there are five categories to define the attention status of a driver: attentive, distracted, looked but didn't see, sleepy (or fatigued), and unknown \cite{stutts2001role}. The attentive category is self-explained.  The ``looked but didn't see'' is usually associated to cognitive distraction of the driver \cite{dong2010driver}. 
Most of the attention criteria defined in the majority of works selected for this SLR in Section~\ref{sec:resul_disc} are related to the distraction and fatigue categories. Thus, it is important to define both more deeply to clarify the discussion of this paper.

The lack of attention of a driver that occurs due to distraction is characterized by her/his choice to involve in a secondary task that is not necessary to guide a vehicle \cite{klauer2006impact}. It happens because of the inattention that, according to \cite{willis2003shorter}, is the failure of paying attention or notice something or some situation. A driver is inattentive when ``at any point in time a driver engages in a secondary task, exhibits symptoms of moderate to severe drowsiness, or looks away from the roadway ahead'' \cite{klauer2006impact}. The goal of an ADAS is to issue an alert to bring back the driver to an attention state.
A loss of attention, when caused by fatigue, can happen due to a combination of conditions such as drowsiness and compromised performance. In this sense, the fatigue detection is more complicated to be performed, since it has different types of definition: physical fatigue, central nervous system fatigue, and mental fatigue. To each of these types there is one or more attention criteria to be considered by the detection system. Examples of attention criteria are yawning and blinking ratio, slow reaction, irritability, confusion, head movement, etc. \cite{regan2011driver}.

The incidence of inattention in drivers can be influenced by external (distractions) or internal (physiological condition) factors \cite{young2007driver}. The latter is related to physiological situation, which can be altered due to fatigue, medication, or any other condition that disturbs the concentration of the driver while conducting the vehicle. The former can include any movement on the streets, pedestrians, vehicular sound systems, people inside the vehicle, cellphones, or other conditions unrelated to the driver. 

Therefore, we will consider the described concepts about attention criteria in the following sections of this paper. More information can be found in \cite{regan2011driver,dong2010driver}

\section{Related Work}
\label{sec:trab_rela}

It is important to notice that among the discussed works we identified only one related SLR \cite{ramzan2019survey}. However, this review does not invalidate the originality of our work, since the scope of the reviews are distinct. While \cite{ramzan2019survey} is interested in reviewing works related to drowsiness detection, we describe more general results about attention criteria detection. 
Now, we present a brief discussion of some published reviews.


The review in \cite{mittal2016head} presents techniques for sleepiness detection of drivers under long hours of driving conditions without rest. The addressed techniques use measurements classified as subjective (e.g., the driver alertness is indicated due to the rate of eye blinking), behavioral (e.g., head movement), psychological (e.g., Electrocardiogram (ECG) data to measure the drowsiness) and vehicular (e.g., driving pattern). This work concludes that obtaining of behavioral metrics are the less invasive to the driver. 
To review the drivers' behavioral detection techniques, e.g., distraction, experience, fatigue, and classify them into techniques as real-time or not, \cite{chhabra2017survey} presents a comparative analysis of advantages, disadvantages and methodologies for intelligent transport systems.
A review in detection systems for drivers' drowsiness is introduced in \cite{kumari2017survey}. Some concepts related to sleepy drivers and factors that lead to sleepiness are defined. The authors briefly describe methods and measurements based on vehicles, behavioral and psychological factors. They conclude that sleepy drivers can be as fatal as drunk drivers.
The review presented in \cite{chowdhury2018sensor} discusses sensors to detect drowsiness and its physiological characteristics, such as Electroencephalogram (EEG) and ECG. The authors also discuss technical aspects, advantages and limitations. The current technologies, monitoring devices, parameters and detection are presented as well.
\cite{ramzan2019survey}  presented a systematic review of methods to drowsiness detection, and also the most used techniques in the reviewed literature. The review is performed in three stages: the classification of techniques based on behavioral, vehicular, and psychological parameters; the supervised techniques with the best performance; and the advantages and disadvantages of each technique. 
A review of recent works in the context of drowsiness detection with deep learning is proposed in \cite{ukwuoma2019deep}. This work presents the five steps of the detection system, composed by the video capture, face detection, characteristic extraction, characteristic analysis, and classification. It also discusses the pros and cons of three approaches of classifications: Support Vector Machine (SVM), hidden Markov model (HMM) and Convolutional Neural Networks (CNN).

\cite{wang2006driver} presents a review with respect to research about fatigue detection, with structural classification of the related work. The studies are mainly focused on measurements of the driver's condition, driver's performance, and a combination of conditions and performance of the driver. Some problems with the approaches were indicated, such as identification of drowsiness condition and the combination of different measurements to obtain better detection results. 
In \cite{coetzer2009driver}, the publication presented a review of methods to drivers' fatigue detection as well. According to the review, the techniques based in EEG measurements are the most reliable. At that time, the authors detected an increase in the use of techniques based on the head movement detection and facial characteristic extraction, besides the computational vision techniques. Finally, the authors recommended the use of hybrid techniques that monitor both the driver and driving pattern to increase the efficiency in fatigue detection.
A review is presented by \cite{dong2010driver} in the context of lack of attention monitoring, by using systems to provide a safer direction. The authors group inattention in fatigue and distraction, which can be categorized in different types and levels. Based on this, the major set of identified techniques in the review focuses on the detection of eyesight distraction, while the minor part focuses on the use of cognitive signals. None of the techniques are focused on auditing nor biomechanics distraction (completing the four distractions defined by National Highway Traffic Safety Administration - NHTSA). The work also provides ideas for future work in the discussed research field.


One more paper, \cite{kaplan2015driver}, discusses well-established techniques in the context of inattention and fatigue monitoring, and also introduces the use of techniques involving mobile devices. Detection methods based on visual characteristics, such as eye blinking and yawning, and non-visual, such as psychological analysis and vehicular parameters, are presented. The authors relate the selected works in a table that describes the most relevant features of each method and even the dissemination of technologies in the context of vehicular companies.
The work in \cite{loce2017driver} presents a review of fatigue detection methods divided in two categories: those that use face imaging cameras and those that, through the vehicle's steering pattern, infer the driver's fatigue level. The approached techniques cover video acquisition, face detection and alignment, eye detection and analysis, head position and eye direction estimation, facial expression analysis, multi-modal sensing and data fusion.

The technical report \cite{ranney2008driver} introduces a review about {drivers' distraction for government uses}, with the aim to help on the elaboration of public policies, regulations and laws. The review mainly discusses the effects of cellphone use, as well as the role of other technological developments, as navigation systems, in the contribution to rise the drivers' inattention. This report also indicates areas of study poorly explored at the time of its publication. 
The focus of the review in \cite{young2007driver} is to discuss the works that consider distraction activities or objects inside the vehicle, such as cellphones and GPS. A discussion about vehicular computational devices that can be designed to reduce the distraction caused by them in drivers is presented.

\section{Review Protocol}
\label{sec:prot_rev}

In order to carry out the proposed SLR, it is necessary to define the review protocol, which characterizes a script that must be followed when executing the review process and involves the definition of the following aspects: a) research question; b) search strategy, which defines the keywords and the search query; c) inclusion and exclusion criteria for primary studies; d) the data to be extracted from the selected primary studies. These steps allow the search for academic papers to be consistent with the considered research problem and, further, to select these articles. By following a well-established protocol, it is possible to carry out a review that minimizes the possibility of selecting or not articles based on the bias of the involved researchers. \cite{kitchenham2004procedures}.

\subsection{Research Question}

Once the need to perform an SLR is identified in a certain area, it is important to define the background or rationale for the review. Thus, it is possible to define one research question or more to be answered.

In the context of this work, the objective is to identify academic work in the literature that can show what has already been proposed as an approach, process, technique, algorithm or solution to the problem of detecting drivers' attention based on images using computer technologies. Therefore, we want to answer the following research question:

\begin{center}
``{Which computational approaches are proposed} in the literature for detecting drivers' attention based on images?''
\end{center}

\subsection{Search Strategy}
Given the research question, it is necessary to define a strategy to search for articles that can give an answer. For that purpose, the terms to be searched are defined and organized to be consulted in the sources of academic publications, including databases, search engines, academic journal sites, congress proceedings, reference section of selected papers, etc.

In this paper, the following terms are used to compose the search for academic papers: \textit{attention, driver, image, detection}. Using the logical operator AND, along with the terms defined above, the following query is defined
\begin{center}
   ``attention'' AND ``driver'' AND ``image'' AND ``detection'' 
\end{center}


Search tools are used to run queries and return academic studies that have metadata related to the search terms defined in the protocol. They are usually made available by institutions that index academic works. In this review, the query is made using the following tools:IEEExplore\footnote{ieeexplore.ieee.org/Xplore/home.jsp} and ACM Digital Library\footnote{dl.acm.org}. These two repositories contain a wide variety of academic works in computing and engineering, as well as in related areas. 



\subsection{Inclusion and Exclusion Criteria}

The search strategy, when applied, returns several articles to be selected or not for review. To make a proper selection, inclusion and exclusion criteria are defined. Based on these criteria, the returned studies will be evaluated and classified as related to the research question or not. These criteria must be chosen in such a way as to be consistent with what has been defined in the review protocol so far, and with the rationale for the review.


Publications that meet the following inclusion criteria are selected:
\begin{enumerate}
    \item the work presents some computational approach for attention detection;
    \item the solution presented may or may not use images to attention detection;
    \item the solution presented may or may not be in the context of automotive driving;
    \item white or gray papers can be selected;
    \item patents.
\end{enumerate}

Publications that satisfy one or more of the following exclusion criteria are not selected:
\begin{enumerate}
    \item works that are not in Portuguese or English;
    \item slide presentations;
    \item entire journals or proceedings must not be considered;
    \item abstracts and extended abstracts.
\end{enumerate}

\subsection{Data Extraction Form}

After the selection of primary studies by applying the inclusion and exclusion criteria, it is necessary to extract the necessary information to carry out, in fact, the discussions and analyses on the proposed theme. Taking into account the research question raised, the data extraction form is constructed. It defines the data or information that must be extracted from the selected works. For this SLR, we want to extract the following information: a) title of the work; b) authors; c) elements of attention used to deduce the driver's level of attention; d) detection technique used, e) sensor used to capture the elements of attention, f) description of the solution presented.

\section{Performing the Review}
\label{sec:exec_rev}

\begin{figure*}[htb!]
    \centering
    \includegraphics[width=0.7\textwidth]{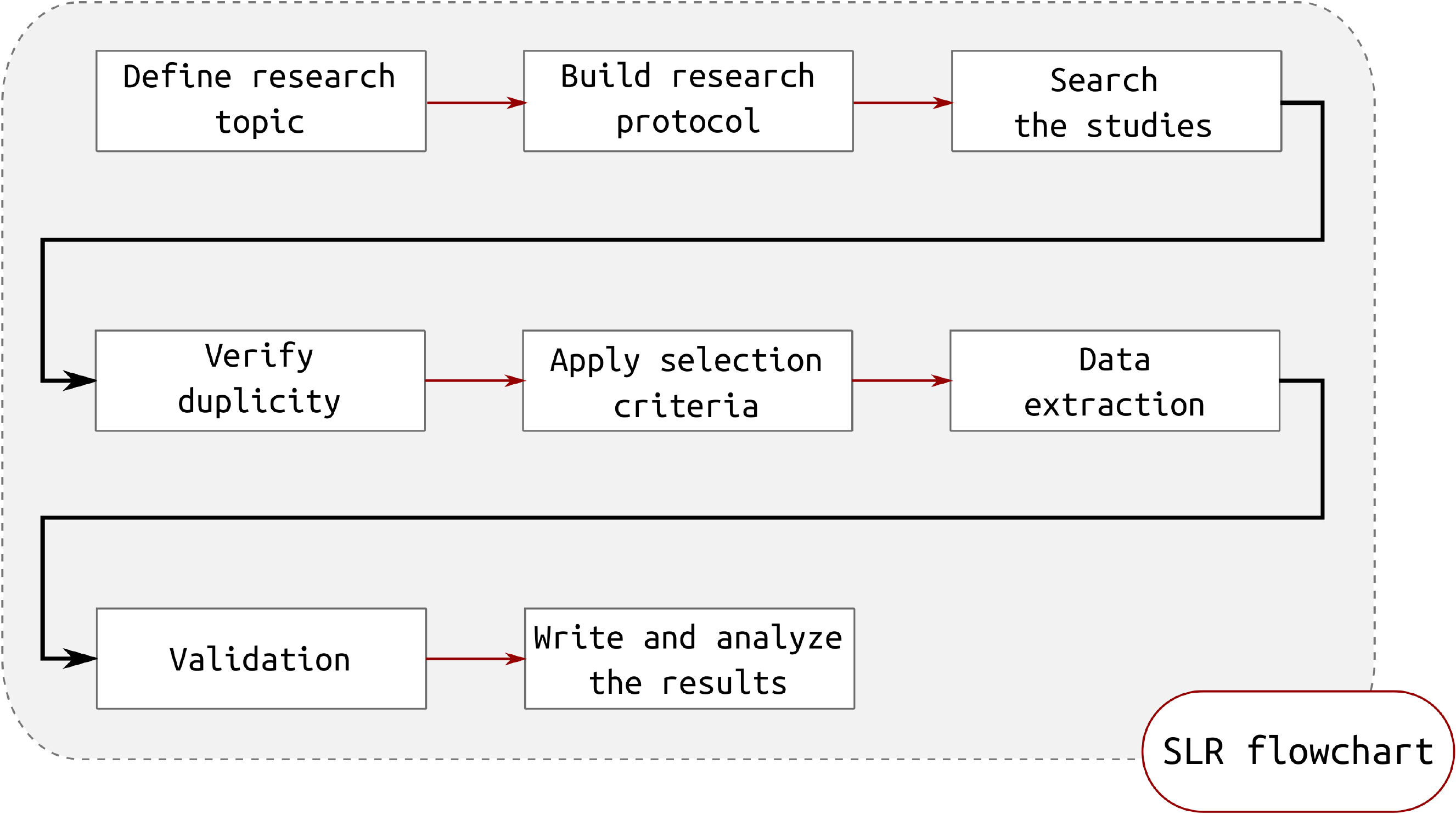}
    \caption{Flowchart of the performed SLR.}
    \label{fig:fluxo_rsl}
\end{figure*}

In general, the SLR consists of the following steps: a) define the research topic; b) build the research protocol; c) search for the primary studies; d) duplication check; e) apply selection criteria; f) data extraction; g) validation; h) write and analyze the results. 
To estimate the quality of the review, at the end of the process, the results are submitted to validation. Fig.~\ref{fig:fluxo_rsl} illustrates the steps to execute the SLR. This section presents the description of the review process from the stage of searching for primary studies. The process was followed by using the guidelines defined in the review protocol.


In the primary studies search stage, the defined query in the review protocol is executed in the search tools. Basically, this step is the collection of academic publications that will be analyzed later applying the inclusion and exclusion criteria. In an attempt to test the query, it returned approximately $1000$ studies.
As the proposed SLR is part of the context of an undergraduate research, this amount would make the review unfeasible. However, we observed that, $80\%$ of the publications belong to the range from 2010 to 2020.

We restrict the review to the most cited studies returned in the query, in order to apprise which methodologies, sensors and/or attention criteria were proposed in them. This procedure could possibly allow the selection of the most relevant approaches in the topic of interest. Moreover, we believe this set of studies could significantly represent the literature, but using a smaller portion of the returned studies.




Thus, in order to reduce the returned quantity but maintaining quality in the selection of primary studies, the review process was applied to $5\%$ of the most cited articles published between 2010 and 2020, plus $5\%$ of the most cited studies in the remaining years of publication, for the search in ACM Digital Library. For the IEEExplore engine, the same criteria were used, except for the percentage of $10\%$. 

Thus, we believe that, with the adopted criteria, this review highlights the most relevant studies of the literature in the area of interest, returning the total of $50$ primary studies, of which $17$ ($34\%$) from IEEExplore and $33$ ($66\%$) from the ACM Digital Library. 


Initially, the duplication check step removes the duplicates of studies returned by search engine. For the selection step, the inclusion and exclusion criteria must be applied to the returned studies. To identify their relationship with the scientific question of SLR and the selection criteria, the study is verified through the analysis of its elements of textual structure. The reading was carried out in the following order: abstract, conclusion, and introduction. This step resulted in the selection of $15$ primary studies. 

In the data extraction step, the selected primary studies have their data extracted according to the form specified in the review protocol. The obtained data are tabulated for further analysis. To streamline the process, this step can be performed along with the application of the selection criteria step. In the last step, this document was developed.

\subsection{Validation}

The validation step consists of giving a subset of returned studies to a group of researchers to apply the same review process defined in the protocol in this subset by applying the selection criteria step. The researchers must be independent and without any knowledge on which studies were selected by the reviewer, thus the results can be cross-checked with no bias. 
At the end of the validation process, the percentage of selected studies matching the reviewer's process and the process performed by the researchers is calculated. 

In this work, the validation step was performed by three researchers. The reviewer created a set of studies according to the following rules:
\begin{itemize}
    \item The set to be given to the researchers will contain $60 \%$ of the total number of articles returned at the duplicity verification stage, as long as the articles selected by the reviewer do not exceed this proportion.
    \item The articles selected by the reviewer should lie within this $60\%$.
\end{itemize}

As a part of the validation, the result of the cross selection was used as a quality metric of the selection performed by the reviewer. In this context, the studies that had divergences regarding the cross-selection were reviewed and discussed in order to understand their inclusion or exclusion. With this dynamic, the percentage of simultaneously selected studies increased, and the validation step became more adequate to the work proposal.

Before we proceed with the validation discussion, consider the following definition: \textit{Let $A$ a finite set. Then, we denote the cardinality of $A$ as $\# A$}. First, we need to define the sets $P$ and $R$, which denote respectively the set of selected studies by the three researchers and the reviewer. The set $P$ is defined by

\begin{equation}
    P = P_1 \cap P_2 \cap P_3
\end{equation}
where $P_1$, $P_2$ and $P_3$ are the sets of selected studies by each of the three researchers. 

\begin{figure}[htb!]
    \centering
    \includegraphics[width=0.4\textwidth]{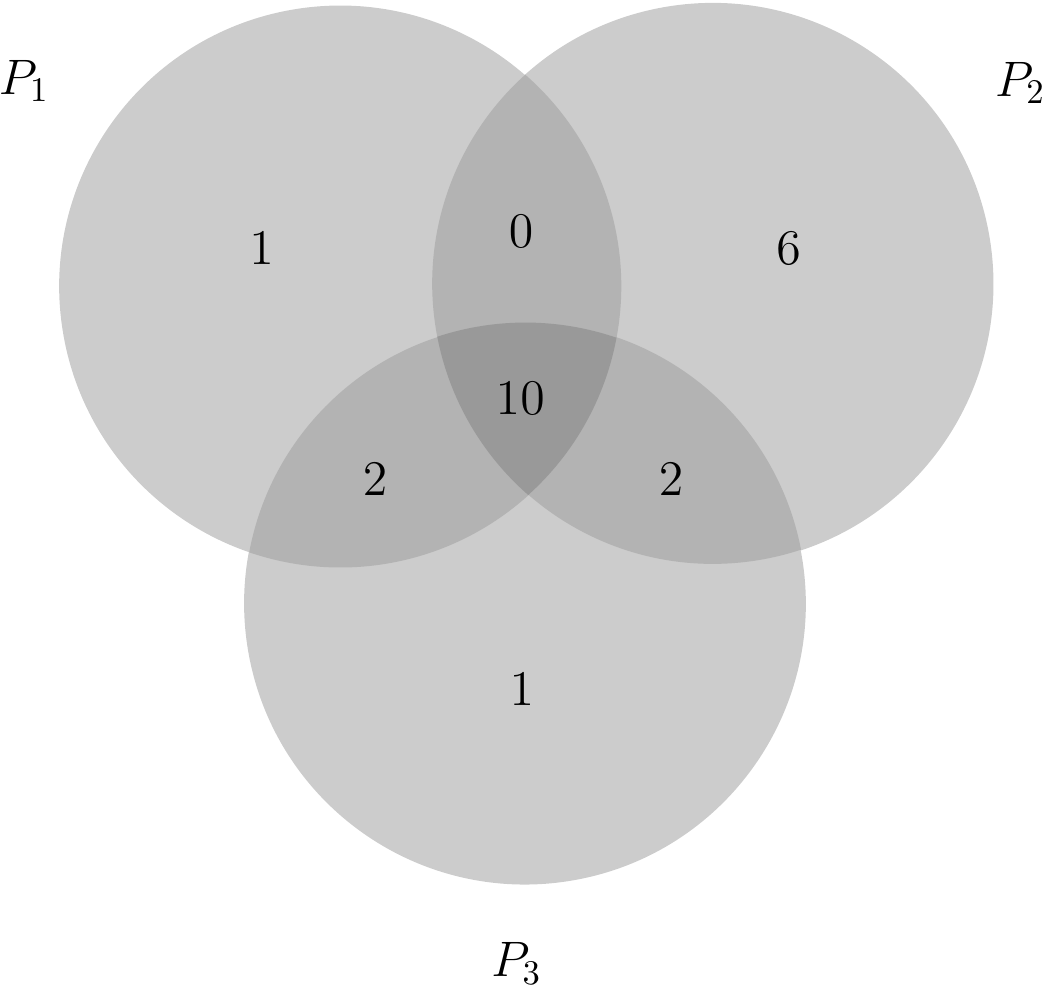}
    \caption{Venn diagram of the sets of selected primary studies by the researchers $P_1$, $P_2$ e $P_3$, their intersections and the set $P$. }
    \label{fig:venn_validacao}
\end{figure}

The reviewer separated a set containing $60\%$ of the total, corresponding to approximately $30$ studies. The quantity of selected articles by each researcher is given by $\# P_1=13$, $\# P_2=18$ and $\# P_3=15$, in which $10$ of the chosen studies matched, i.e., $\# P=10$, as shown in Fig.~\ref{fig:venn_validacao} by a Venn diagram.

For associated the percentage $P_v$, we define the equation below.
\begin{equation}
    P_{v} = \frac{\#(P \cap R)}{\#R}*100
\end{equation}

\begin{figure}[H]
    \centering
    \includegraphics[width=0.35\textwidth]{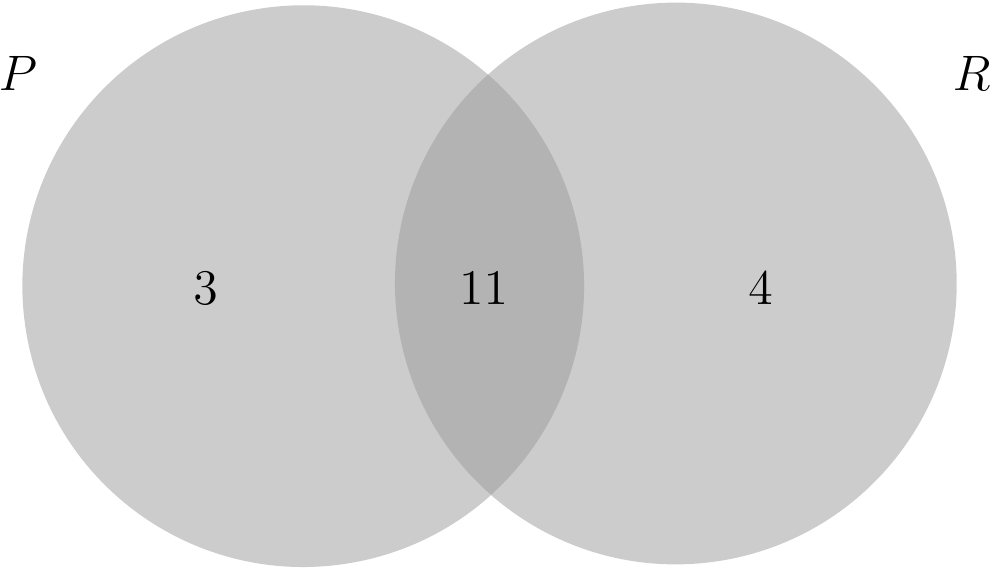}
    \caption{Venn diagram of the sets of primary studies $P$ and $R$ e their intersection.}
    \label{fig:val_result}
\end{figure}

We obtained $P_{v} = 73,33\%$ for the validation process. Fig.~\ref{fig:val_result} shows the Venn diagram of the selected studies by the researchers, reviewer and their intersection.

The obtained percentage $P_{v} = 73,33\%$ is considered as an improvement indicator for the SLR process. The validation results allowed the revaluation of which studies should be kept, excluded or added in the final results of SLR. After these considerations, the post validation presented a selection of $22$ primary studies, which is the number of studies the reviewer extracted the information accordingly to the data form defined in the review protocol. We consider that the process of validation allowed the SLR to achieve an improvement on its quality in terms of obtaining a set of studies better related to the defined research question. 

\section{Results and Discussion}
\label{sec:resul_disc}
In this section, we present a discussion of the selected primary studies and their descriptive statistics to conclude the SLR process in driver's attention detection, answering our research question ``Which are the computational approaches proposed in the literature for detecting drivers' attention based on image?''. We organize the extracted information and also discuss the relation between the criteria and the selected studies in order to provide a useful resource to the interested reader.

\subsection{Selected Primary Studies}
Herein we present the extracted data from selected primary studies in Table~\ref{tab:rsl_resultado}. We found it interesting to sort the articles by publication year, from the oldest to the most recent, to provide a chronological view of proposed solutions. A discussion of each article can be found below.

  \begin{center}
    \begin{longtable}{|p{1.2cm}|p{0.9cm}|p{4cm}|p{4cm}|p{3.6cm}|}
    \caption{Extracted data from primary studies selected in the SLR.}\label{tab:rsl_resultado}\\
    \hline
    \multirow{2}{*}{\textbf{Article}} & \multirow{2}{*}{\textbf{Year}}  & \textbf{Employed} & \textbf{Attention} & \multirow{2}{*}{\textbf{Sensor}} \\
    &  & \textbf{Techniques}  & \textbf{Elements} &\\
    \hline
\cite{sodhi2002road} & 2002	& 	Graphical analysis of the position of the eyes along the direction. & Movement of eyes and where they are focused & A device called HEAD, with three cameras: one for the environment, one for the eyes and an infrared camera.\\
\hline
\cite{smith2003determining}  & 2003 & Finite State Automata (FSM)	& 	Eyes and lips monitoring, and metrics related to their attributes. Delimitation of the face region.		& Unique camera.\\
\hline
\cite{mccall2006driver}	& 2006 & 	Particle filter; Dynamic Bayesian network.	& 	Eyes state, eyebrows position.	& 	Camera.\\
\hline
\cite{bao2007projection} & 2007 	& 	PERCLOS; PERROTAT	& 	Eye state and face orientation.	& 	Monocular camera.\\
\hline
\cite{benoit2009multimodal}	& 2009 & 	MPT toolbox, Fourier analysis, data fusion and data fission. & 	Eye state, head position and yawn.	& 	MPT face detector\\
\hline
\cite{zeng2010driver} & 2010	& 	Scale Invariant Feature Transform (SIFT)	& 	Only open eyes can be detected. Attention budget.	& 	Camera.\\
\cite{ye2012detecting} & 2012 	& 	Decision tree.	& 	Gaze direction.	& 	Gaze monitoring glasses and video.\\
\hline
\cite{you2013carsafe} & 2013 	& 	Computational vision; Support Vector Machine (SVM); Hough-line transformation; RANdom SAmple Consensus (RANSAC) algorithm; Decision Tree. & Head position, gaze direction and blinking index. & Smartphone cameras.\\
\hline
\cite{mbouna2013visual} & 2013 	& 	AdaBoost; facial-feature-matching.	& 	Head position, eye and pupil states.	& 	Unique camera.\\
\hline
\cite{dwivedi2014drowsy} & 2014	& 	Multi-layer convolutional Neural Network;  Viola–Jones Features.		& Face	& 	Camera (data set - 30 videos)\\
\hline
\cite{rezaei2014look} & 2014	& 	Asymmetric Appearance Model (AM); Fuzzy logic.	& 	Yawn and head movement.	& Two monocular cameras.\\
\hline
\cite{wang20143d} & 2014	& 	Iterative Closest Point, particle filter.		& Head position & RGB and infrared cameras.\\
\hline
\cite{han2015neuromorphic} & 2015 	& 	Neuromorphic Visual Processing.		& Central point between the driver's eyes.	& 	Camera.\\
\hline
\cite{mihai2015using} & 2015 	& 	Uninformed.	& 	Head position, fixed gaze and eyes state.	& 	Smartphone's front and rear cameras.\\
\hline
\cite{kim2015monitoring} & 2015 	& 	Pose Estimate Algorithm (POSIT); Gestalt Saliency Map Model.	&  Head position from face center detection, eyes, nose and mouth. 	& 	Two monocular infrared cameras.\\
\hline
\cite{zhang2015real} & 2015	& 	Weber local binary pattern (WLBP); Support Vector Machines (SVMs).	& 	Eye state and head position.	&  RGB-D and RGB cameras.\\
\hline
\cite{jha2016analyzing}	& 2016 	& Head pose estimation algorithm (HPA); Linear regression.	&  Head pose and eyes movement.	& AprilTags and glasses with laser pointer\\
\hline
\cite{haq2016eye} & 2016 	& 	Viola-Jones Features. & Eye blinking number.	& 	Camera.\\
\hline
\cite{wang2017head}	& 2017 	& Pose from Orthography and Scaling with ITerations (POSIT)	& Head and eyes position. & Camera.\\
\hline
\cite{husen2017syntactic} & 2017 	& 	Syntactic pattern recognition technique.	& 	Visual attention, velocity, wheel angle and signals of Atenção visual, velocidade, ângulo da roda and lane change indication signs.	& 	Three cameras e two IR sensors.\\
\hline
\cite{deCastro2018distraction} & 2018 & K-nearest Neighbor Classifier (k-NN) & Eye gaze, head pose and lips. & Camera.\\
\hline
\cite{huang2019information}	& 2019 	& Multi-Stream Neural Network; Multi-Task Cascaded Convolutional Networks.	& 	Eye estate and mouth.	& 	Uninformed. Only presents the classifier training with YawDD dataset. \\
\hline
    \end{longtable}
\end{center}

In \cite{sodhi2002road}, the authors propose an attention detection system that uses a detection device called HED (head mounted eye tracking device). It analyzes attention from the movement of the eyes and the direction of the gaze, captured in ASCII and MPEG files, from where the gaze position is extracted at a given time. The horizontal and vertical positions of the eyes along the direction are plotted, and the graphic patterns indicates whether the driver is paying attention or not in the driving task. An important piece of information is that the proposed system needs calibration for each user.

Through video analysis (frames), the system proposed in \cite{smith2003determining} detects face rotation in all directions. It also detects mouth/eyes concealment, closing, blinking, and yawning. The solution represents the eye look fixation in 3D to check the direction in which the driver is paying attention. This system is activated even when the face is hidden due to head rotation and its mechanism for detecting the driver's attention is modeled as finite state automaton.

In \cite{mccall2006driver}, a multilevel steering assistance system is proposed to first identify the driver's head and then the facial regions. Facial features, such as eyebrows and the corners of the eyes, are also analyzed. In this system, a classifier based on a Bayesian Dynamic Network is employed and trained based on naturalistic data, i.e., those obtained in real-world situations. According to the authors, this allows the creation of classifiers that infer more accurately the intention and attention of the driver when compared to the use of training data created in the laboratory or simulations.

The work in \cite{bao2007projection} proposes a low-cost detection system, based on a monocular camera to monitor driver fatigue. The system uses two metrics: Percentage eye openness tracking (PERCLOS), which monitors the percentage of eyes closed over time, and PERROTAT, which calculates the percentage of head rotations. These metrics are used to estimate the driver's status from the face analysis. This system also monitors the driver's attention through the opening/closing of the eyes and the position of the head, through the center of the mouth and the distance between the eyes. 

\cite{benoit2009multimodal} proposed a driving simulator that uses a driver status detection system based on video data and biological signals. The simulator analyzes, along with the algorithm, the user's mental state, using data on stress levels (by cardiac monitoring), eye and mouth detection, yawning, closing eyes and head rotation. A fusion of the attributes obtained to detect hypo-surveillance in the steering is performed.

A methodology that embarks on an automatic co-pilot system to assist in driving a vehicle is proposed in \cite{zeng2010driver}. In the presented solution, active and passive sensors are used to define the state for vision-based vehicle detection. The system uses face monitoring and, after fixing itself in position, can identify the eyes and whether they are open or not. In order to increase detection efficiency, a slight tolerance for head rotation is considered. The strategy used by the solution to measure distraction is known as attention budget. It considers that each driver has a set of distracting characteristics according to personal features, such as age, medical condition, time of day, etc. The strategy used stays in the background when eyes are off-track or closed.

In \cite{ye2012detecting}, a system to detect the eye contact between an adult and a child is developed in order to check the child's attention. The system verifies the adult's point of view through glasses and the direction of the child's eye through a computer vision analysis of the video. The gaze and face information for each frame in the video is used to train a classifier that detects the existence of eye contact in a particular frame. The selected study objects are used in order to be applied to medical, therapy, classroom and childcare devices.

To analyze the behavior of the driver and the driving pattern on the road, a methodology is proposed in \cite{you2013carsafe} to analyze the information from the front and back cameras of a cellphone simultaneously. Once this is done, the data from both cameras are crossed to obtain a response of attention or inattention from the driver.

Unlike most studies that check if the driver eyes are closed and the angle of the driver face, \cite{mbouna2013visual} proposed a visual analysis of the state of the eyes and the position of the head, involving concepts such as the eye index and pupil activity to extract critical information from drivers' lack of attention. In the presented method there is no need to use additional sources of light beyond that provided by the system. The experiments described show that this type of approach can help to achieve a better performance in detecting sleepy or distracted drivers.

In \cite{dwivedi2014drowsy}, the authors propose a methodology based on the extraction of visual characteristics from data patterns, without human intervention. This is done with a deep learning model, using convolutional neural networks. The feature maps produced by the convolutions are used to detect driver drowsiness. A soft-max classifying layer is also used to classify the extracted frames.

An asymmetric appearance model, 2D or 3D, to estimate the position of the head using a Fermat-point transform and an adaptation of Global Haar is presented in \cite{rezaei2014look}. In this study, the performance is analyzed in real time for driving scenarios in real world, taking into account a normalized object of the driver's face together with components of the road. This system defines the driver's attention, yawning, head movement, vehicle detection and distance calculation.

A 3D head position indicator is used for an attention detection solution presented in \cite{wang20143d}. By defining and calibrating certain gaze zones in rear-view mirrors of the vehicle, a template is defined for the driver to be aware of. The head position is captured by an infrared camera for attention analysis. The 3D point cloud of the head is generated and used to calculate the driver's head rotation. Finally, the template is used, comparing the points of the cloud of the last position of the head with the points of the cloud referring to the current position.

The use of a neuromorphic processing algorithm based on a biological vision system to detect people from a moving vehicle, which may be the driver or also vulnerable people on the track, is presented in \cite{han2015neuromorphic}. The results obtained have a detection rate of $99\%$ by day and $88\%$ by night. The proposed system is fast and robust and can be embedded in Field Programmable Gate Array (FPGA).

An application called NAVIIEYES, an ADAS based on smartphone cameras, is proposed in \cite{mihai2015using}. This application analyzes the driver's attention status and the vehicle's traffic environment, in order to warn the driver about two to three seconds before any possible impact with obstacles. The study classifies attention into two types: drowsiness and driver behavior.

In \cite{kim2015monitoring}, an ADAS based on a technique called Head Pose Estimation 3D is proposed to estimate the driver's attention area. Two analyses are performed to detect attention: internal, to check the driver's head pose; and external, from the overhang map (Gestalt saliency map). The generated data are combined to estimate the amount of information the driver is being subjected to at any given time. The driver's head orientation is detected by a POSIT algorithm. Given the position, it is checked whether the head is within the area of attention or not.
 
The use of RGB-D cameras (coupled to a Kinect motion detection device) in a driver fatigue detection system is proposed in \cite{zhang2015real}. This type of camera provides extra depth compared to conventional RGB cameras. The data generated by RGB-D and RGB cameras are combined to generate information for analyzing the driver's head and eye positions.

In \cite{jha2016analyzing}, the authors explore detection models that consider the interference of the driver's head position and attention state. The position of the "target" is analyzed, while focusing on predefined markers on the vehicle's mirrors such as side windows, speed panel, radio, etc. Linear regression models for detection are proposed, which are effective in predicting the location of the gaze, especially in the horizontal direction. The study also deals with the bias introduced by the movement of the eyes during driving and the position of the head with the gaze directed to certain areas. The article is interested in analyzing the relationship between these two criteria.

The level of attention is determined from the driver's blink rate in \cite{haq2016eye}. It is known that a person's level of attention can be estimated from the blink rate. Therefore, depending on the person's condition, the blink pattern may vary. The proposed methodology, when applied to vehicles, helps considerably in reducing the number of accidents related to fatigue, as it issues a warning to the driver, that can take preventive measures. The evaluation of the blink rate is defined from the data capture and a threshold.

The study in \cite{wang2017head} proposes an appearance-based head pose-free eye gaze prediction method to estimate the driver's gaze zone considering the free movement of the head, which, according to the authors, goes against traditional methods based on eye models. The proposed method is applicable in a real driving environment.

The study in \cite{husen2017syntactic} analyzes the pattern of vehicle changing lanes through the driver's speed, wheel angle and focus, with the aim of ensuring that the driver is driving safely. The approach prevents accidents that may occur due to sudden changes in lane. To detect the pattern of track changes, a syntactic pattern recognition technique is employed to allow the generation of structural information defined by the grammar. It is used for string sequences, which are analyzed by a parser that interprets the behavior of the driver on the wheel.

In \cite{deCastro2018distraction}, a driver attention detection model based on visual distraction and mouth movement is presented. The authors argue that the main factors that cause traffic accidents are visual distraction and passenger conversations with the driver. To perform the detection of attention, initially, the position of the head and features of the mouth are extracted from videos captured from the driver. Then the frames are analyzed by a binary classification (0 for not distracted and 1 for distracted), where the k-Nearest Neighbor Classifier (k-NN) is used. In order to increase the accuracy of the detection, the attention indication elements are also checked for time intervals (Time Restriction Algorithm). The model is validated with the K-Fold method with $95\%$ of detection power.

Two individually trained CNNs are used in \cite{huang2019information}, each of which receives input from each eye. Then, the information generated by the networks is crossed into an interaction module. In the pre-processing step, the driver's facial limits are obtained and five landmark points (position of the right eye and left eye, left and right nose and lips). The authors use the ZJU Eye blink Database and the Yawn Detection Dataset (YawnDD).

\subsection{Descriptive Statistics of the Selected Primary Studies}

In Table~\ref{tab:rsl_resultado}, we notice that employing machine learning techniques in attention detection problems started in $2010$. It can be related to the popularization of this research field at the same period. We also found that the majority of the solutions presented uses more than one technique to the detection process, which indicates that there is not an isolated method that solves the problem, and the planning of the detection solutions must consider different approaches in different stages of the solution in order to improve their performances.

\begin{figure*}[h]
    \centering
    \includegraphics[width=0.8\textwidth]{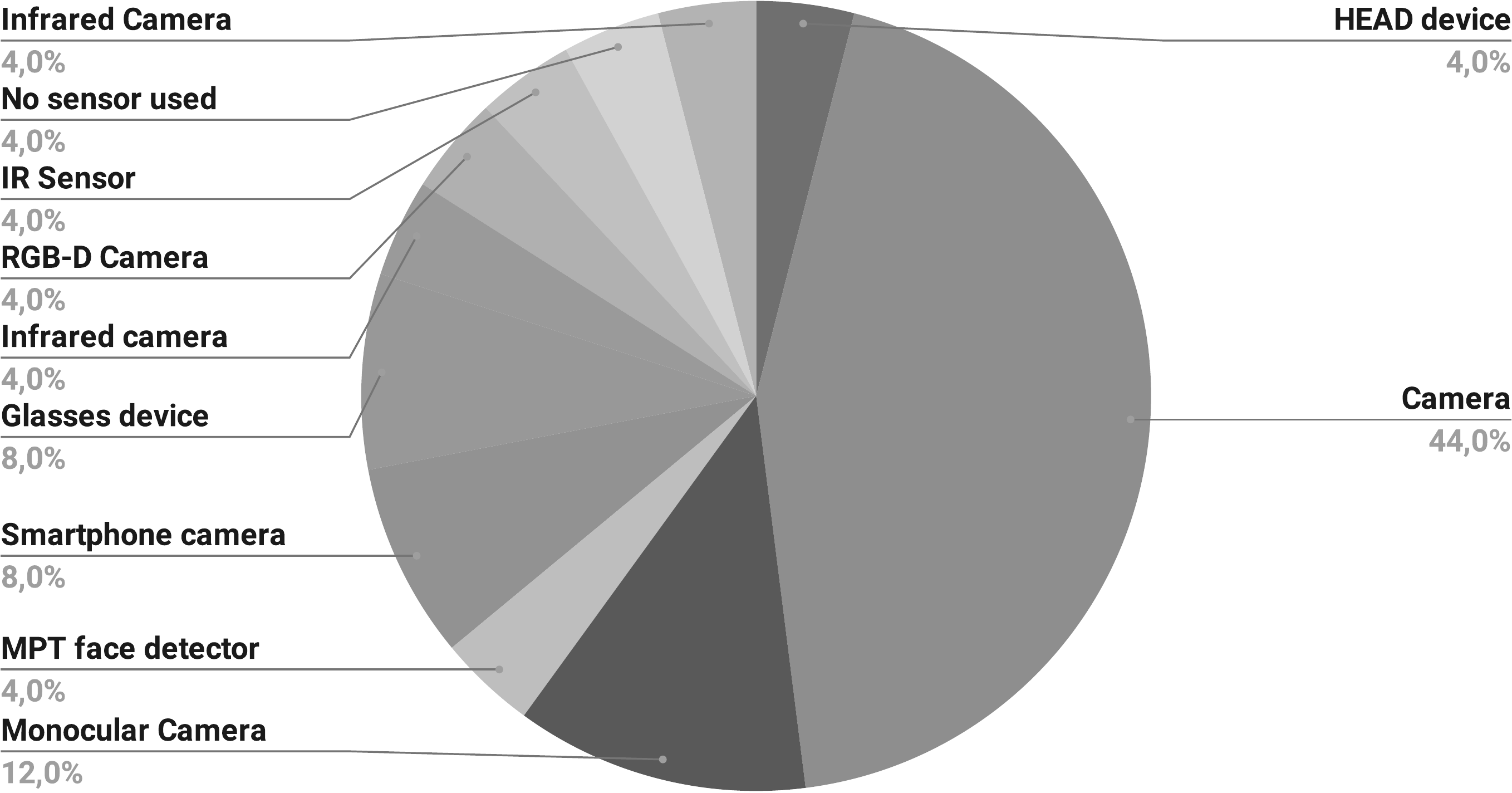}
    \caption{Sensors used by the selected studies in SLR.}
    \label{fig:chart_sensors}
\end{figure*}

Fig.~\ref{fig:chart_sensors} shows the most used detection approaches in the selected studies. In most of them it was specified, without any further technical details, that cameras were used to capture the elements of attention. Due to the absence of technical details, it was not possible to generate classes with greater granularity in this category of sensors. Then, monocular cameras were used in $12\%$ of the primary studies. The other types of sensors presented a percentage of $4\%$ each, except smartphone cameras that were used in $8\%$ of the proposed solutions. It is interesting to note that the most recent study using smartphones dates from $2015$. With the current hardware embedded in these devices (powerful and increasingly cheaper), we expected that we could find them in more recent paper.

A frequency matrix of the attention detection techniques is presented in Table~\ref{tab:freq_matrix}. We consider the attention elements found in the selected studies. When a technique is involved in the solution to monitor an attention element, it is denoted by the number $1$ in the table. The results suggest there is a great variety of employed techniques in the context of this review, and that a consensus of which combination of techniques or elements ideal to solve the problem of attention detection could be difficult. 


We also observe, in Table~\ref{tab:freq_matrix}, that the most used attention element is the head position, which is cited 21 times in the Table. This criterion can be considered similar to face orientation, but its frequency of citation is organized as a distinct class due to the manner used to reference it in the selected studies. 
Subsequently, we have the eyes state with 18 citations. Many of the selected works use two elements in the same solution to increase the accuracy of the attention detection. The less used elements are the driving pattern, eyebrows, lips states, and mouth state (depending on the extracted features, these two can be considered as equivalent). A suggestion of a possible study to be conducted is verify how differently the less used attention elements could influence on the attention detection in comparison with the most appointed.

 
 \begin{center}
    \begin{longtabu} to \textwidth {|p{2.0cm}|c|c|c|c|c|c|c|c|}
         \caption{Frequency matrix relating techniques with attention elements to detect attention.} \label{tab:freq_matrix}\\
        \hline
        \textbf{Used} & \textbf{Drive} & \textbf{Eyebrows} & \textbf{Eyes} & \textbf{Face} & \textbf{Head} & \textbf{Lips}	& \textbf{Eyes} & \textbf{Mouth}\\
        \textbf{Techniques} & \textbf{pattern} & \textbf{state} & \textbf{state} & \textbf{orientation} & \textbf{position} & \textbf{state}	& \textbf{gaze} & \textbf{state}\\
        \hline
        AdaBoost&&&1&&1&&&\\
        \hline
        Asymmetric Appearance Model (AM)&&&&&1&&&\\
        \hline
        Computer vision&&&&&1&&1&\\
        \hline
        Decision Tree&&&&&1&&1&\\
        \hline
        Dynamic Bayesian Network&&1&1&&&&&\\
        \hline
        Facial-Feature-Matching&&&1&&1&&&\\
        \hline
        Finite State Automata (FSM)&&&1&&&1&&\\
        \hline
        Fourier Analysis&&&1&&1&&&\\
        \hline
        Fuzzy Logic&&&&&1&&&\\
        \hline
        Gestalt Saliency Map Model&&&&&1&&&\\
        \hline
        Graphical analysis&&&1&&&&&\\
        \hline
        Head Pose Estimation Algorithm (HPA)&&&1&&1&&&\\
        \hline
        Hough-line transformation&&&&&1&&1&\\
        \hline
        Iterative Closest Point&&&&&1&&&\\
        \hline
        k-Nearest Neighbor Classifier&&&&&1&1&1&\\
        \hline
        Linear Regression&&&1&&1&&&\\
        \hline
        Multi Layers Convolutional Neural Network&&&&1&&&&\\
        \hline
        Multi-Stream Neural Network&&&1&&&&&1\\
        \hline
        Multi-Task Cascaded Convolutional Networks&&&1&&&&&1\\
        \hline
        Neuromorphic Visual Processing&&&&&1&&&\\
        \hline
        Particle Filter&&&&&1&&&\\
        \hline
        Particles Filter&&1&1&&&&&\\
        \hline
        PERCLOS&&&1&1&&&&\\
        \hline
        PERROTAT&&&1&1&&&&\\
        \hline
        Pose Estimate Algorithm (POSIT)&&&&&1&&&\\
        \hline
        Pose from Orthography and Scaling with ITerations (POSIT)&&&1&&1&&&\\
        \hline
        RANdom SAmple Consensus (RANSAC) algorithm&&&&&1&&1&\\
        \hline
        Scale Invariant Feature Transform (SIFT)&&&1&&&&&\\
        \hline
        Support Vector Machine (SVM)&&&1&&1&&1&\\
        \hline
        Syntactic Pattern Recognition Technique&1&&&&&&1&\\
        \hline
        Viola-Jones Feature&&&1&1&1&&&\\
        \hline
        Weber Local Binary Pattern (WLBP)&&&1&&1&&&\\
        \hline
        \textbf{Total}&1&2&18&4&21&2&7&2\\
        \hline
    \end{longtabu}   
\end{center}

With the presented information, based on the extracted data of the selected primary studies in this SLR, we can observe there are gaps in the understanding of which techniques and criteria are the most adequate to driver's attention detection, due to the big number of employed techniques. A deeper understanding about the efficiency of the methods seems to be also necessary. Research that involves solutions with new combination of the attention elements, or even with all the identified elements in SLR could also be conducted.

\section{Conclusion}
\label{sec:conclusion}

The planning, execution description and results of a Systematic Literature Review about driver's attention detection based on image were described in this work. We produced a review protocol that includes the participation of a reviewer and three more researchers and contains the review guidelines. During the SLR process, 50 studies were returned by the search tools, out of which 22 were selected as primary studies related to the research question. 

Initially, the concepts and fundamental definitions about attention, in a general and driving context, were discussed. Then, other existing reviews and surveys were presented and discussed. We also described the search protocol in detail to enable the guidelines verification on which the SLR is based.

This SLR was performed in 8 steps: a) define the research topic; b) build the research protocol; c) search for the primary studies; d) duplication check; e) apply selection criteria; f) data extraction; g) validation; h) write and analyze the results. From each primary study selected, we extracted the data on the year of publication, title, authors, techniques employed, detection strategies, detection criteria and sensors used to capture the criteria. Thus, it was possible to present statistics of the most used technologies (in its broadest sense), among other information.

The presented results can be used as a resource to compose new research projects about driver's attention detection. The extracted data in SLR can also be used as a resource tool on which methods and attention criteria are practicable in order to be used in the development of an ADAS. As a future goal, we would suggest the verification of detection techniques that are viable to be embedded in a prototype built in Single Board Computing (SBC), Computer on Module (CoM) and/or Cloud Computing.

\section*{Acknowledgement}
We thank Dr. Nandamudi L. Vijaykumar (National Institute of Spatial Research - INPE) for his support and orientations on editing and improvements of this Systematic Literature Review.

\bibliographystyle{plainurl}
\bibliography{bibliografia}

\end{document}